\documentclass[10pt]{article} % For LaTeX2e
\usepackage[preprint]{tmlr}

\usepackage{amsmath,amsfonts,bm}

\def\eqref#1{equation~\ref{#1}}
\def\1{\bm{1}}

\DeclareMathAlphabet{\mathsfit}{\encodingdefault}{\sfdefault}{m}{sl}
\SetMathAlphabet{\mathsfit}{bold}{\encodingdefault}{\sfdefault}{bx}{n}

\usepackage{hyperref}
\usepackage{url}
\usepackage{graphicx}
\usepackage{algorithm}
\usepackage{algorithmic}
\usepackage{amsfonts} 
\usepackage{comment}
\usepackage{amsmath}
\usepackage{booktabs}
\usepackage{adjustbox}
\usepackage{subcaption}
\usepackage{geometry}
\usepackage{braket}

\usepackage{amssymb,mathtools}
\usepackage{booktabs} 
\usepackage{adjustbox}
\usepackage{amsmath}
\usepackage{multirow}
\usepackage{booktabs}
\usepackage{braket}

 \usepackage{colortbl}
\usepackage{tcolorbox}
\usepackage{tabularx}
\usepackage{array}
\usepackage{colortbl}
\tcbuselibrary{skins}

\newcolumntype{Y}{>{\raggedleft\arraybackslash}X}

\tcbset{tab1/.style={fonttitle=\bfseries\large,fontupper=\normalsize\sffamily,
colback=yellow!10!white,colframe=red!75!black,colbacktitle=green!40!white,
coltitle=black,center title,freelance,frame code={
\foreach \n in {north east,north west,south east,south west}
{\path [fill=green!75!black] (interior.\n) circle (3mm); };},}}

\tcbset{tab2/.style={enhanced,fonttitle=\bfseries,fontupper=\normalsize\sffamily,
colback=yellow!10!white,colframe=green!50!black,colbacktitle=green!40!white,
coltitle=black,center title}}

\title{\huge Obtaining Favorable Layouts for Multiple\\Object Generation}

\author{\name Barak Battash \email  \\
      \addr Faculty of Engineering, Bar Ilan University
      \AND
      \name Amit Rozner \email  \\
      \addr Faculty of Engineering, Bar Ilan University 
      \AND
      \name Lior Wolf \email  \\
      \addr School of Computer Science, Tel Aviv University
      \AND
      \name Ofir Lindenbaum \email \\
      \addr Faculty of Engineering, Bar Ilan University
}

\def\month{02}  % Insert correct month for camera-ready version
\def\year{2024} % Insert correct year for camera-ready version
\def\openreview{\url{https://https://openreview.net/forum?id=O9RUANpPmb}} % Insert correct link to OpenReview for camera-ready version

\begin{document}

\maketitle

\begin{abstract}
  Large-scale text-to-image models that can generate high-quality and diverse images based on textual prompts have shown remarkable success. These models aim ultimately to create complex scenes, and addressing the challenge of multi-subject generation is a critical step towards this goal. However, the existing state-of-the-art diffusion models face difficulty when generating images that involve multiple subjects. When presented with a prompt containing more than one subject, these models may omit some subjects or merge them together. To address this challenge, we propose a novel approach based on a guiding principle. We allow the diffusion model to initially propose a layout, and then we rearrange the layout grid. This is achieved by enforcing cross-attention maps (XAMs) to adhere to proposed masks and by migrating pixels from latent maps to new locations determined by us. We introduce new loss terms aimed at reducing XAM entropy for clearer spatial definition of subjects, reduce the overlap between XAMs, and ensure that XAMs align with their respective masks. We contrast our approach with several alternative methods and show that it more faithfully captures the desired concepts across a variety of text prompts.
\end{abstract}

\begin{figure}[t]
    \centering
    \includegraphics[width=0.9799\linewidth]{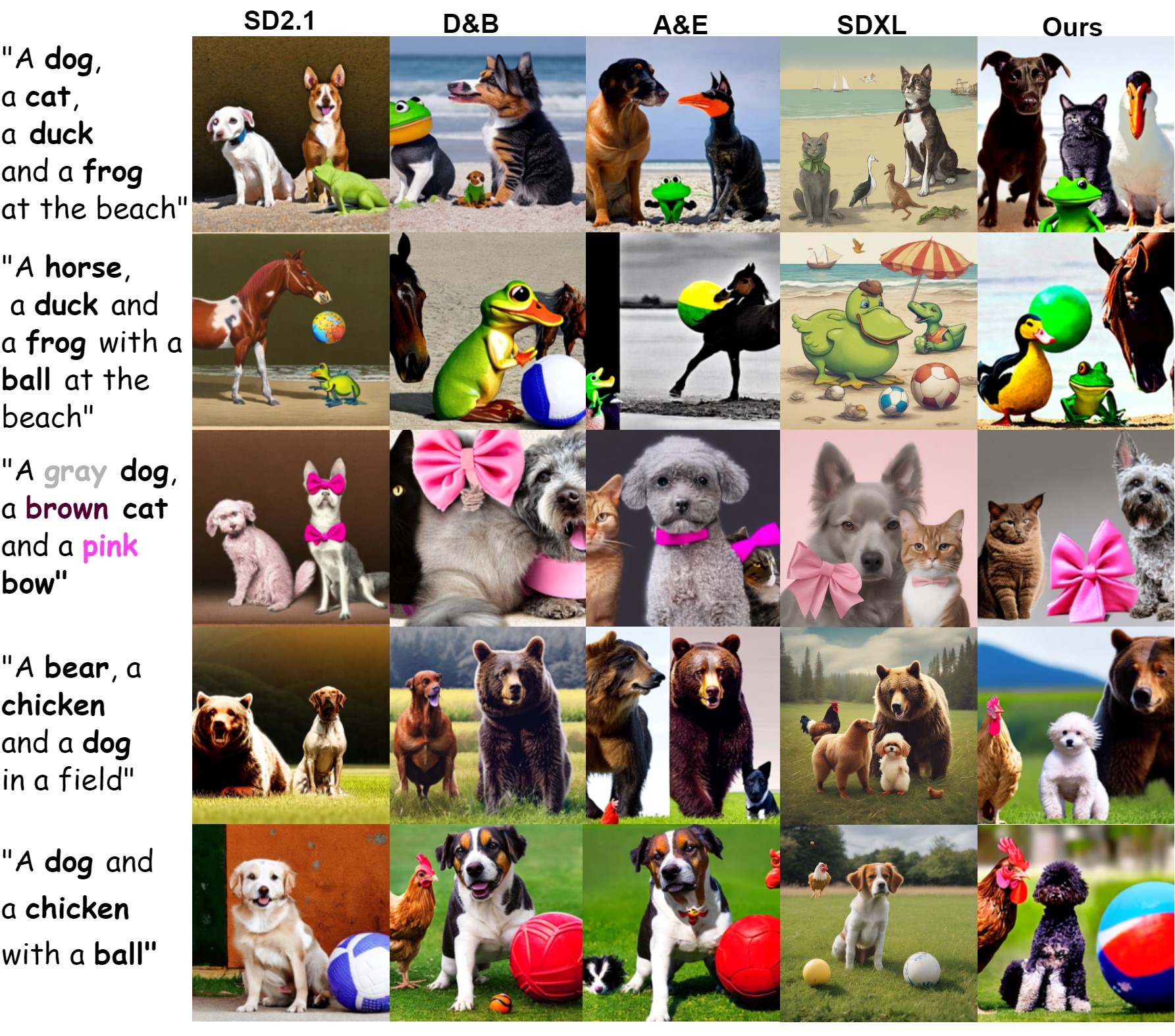}
  \vspace{-0cm}
    \caption{This Figure shows the generation outputs of our method and competitive methods for multiple prompts, with various amount of subjects and objects.}\label{fig:opennew}
\end{figure}
\section{Introduction}
\label{sec:intro}
In recent years, diffusion probabilistic models have garnered considerable attention from researchers across both academia and industry due to their remarkable performance and applicability to a wide range of downstream tasks related to high-quality image generation. State-of-the-art text-to-image foundation models, such as Stable Diffusion~\cite{rombach2022high}, Imagen~\cite{saharia2022photorealistic}, and DALL-E~\cite{parmar2023zero}, are predominantly based on diffusion models.

Diffusion models operate by iteratively denoising a noise-perturbed input image. Notably, the Stable Diffusion model \cite{rombach2022high} has showcased impressive capabilities in generating diverse and realistic images, underscoring the efficacy of diffusion-based approaches in image synthesis tasks.

There are, however, a few eminent semantic challenges within such text-to-image synthesis, three of which are ``subject neglect'', ``subject blending'' and ``attribute binding''. The first pertains to generated samples where not all subjects described in the input text are present in the resulting image. The second is where the subjects are blended into other subjects, for example a bear with elephant trunk. ``Attribute binding'' is a compositional issue, wherein attribute details, such as color or texture, are either misaligned with their intended objects or mistakenly associated with unrelated objects.

This work aims to address the challenges of subject neglect and subject blending, which become more complicated as the number of subjects increases, see Figure ~\ref{fig:opennew}. Although we do not attempt to solve the problem of attributed binding, our approach can be combined with any other attribute-binding method and to show better or compatible results.

Well-structured layout is required to avoid subject neglect and blending. If the layout is such that the subjects are well-separated and no subject dominates the scene, the diffusion methods can generate desirable images. Our research hypothesis is that given an initial noise map $z_T$ the diffusion model has bias towards some favorable layouts. Thus, manipulating the latent map is important as manipulating the attention maps. 

The layout is determined in approximately the initial 15 diffusion steps, but is not apparent during these steps. To mitigate this, our proposed solution has three phases. During the first few denoising steps, we apply various loss terms to encourage the XAMs associated with the tokens of each subject to be excited but separate from other subjects. Our solution to this is considerably more elaborate than previous attempts since we rely on the spatial structure and not just on the maximal activation. 

Then, after these initial steps, we extract binary masks per object and revise these masks to obtain favorable layouts. This is done by shifting some of the objects. The latent space of the diffusion model is then readjusted to match the optimized masks. 

Finally, diffusion continues, but the process is driven such that the per-subject XAMs match the set of fixed masks we have previously generated. Overall, the method provides a comprehensive solution to the challenges of multi-subject generation across all diffusion steps, all subjects, and the various spatial locations.

In an extensive set of experiments, we demonstrate that our proposed method outperforms in all established metrics the many baselines that exist in this field by a sizable margin. Quantitative results demonstrate that these qualitative advantages match the obtained visual improvement. Figure  \ref{fig:opennew} presents several examples of images generated using competitive methods. Widely used text-to-image diffusion based generative models struggle to generate an image containing all the objects in the prompt. Our model improves the image layout and leads to generated images that are more faithful to the prompt.

% \begin{figure}[t]
%     \centering
   
%     \includegraphics[width=1.02\linewidth]{gallery/ECCV.drawio.png}
%     \caption{We present a training-free procedure to automatically re-arrange the layout of attention maps in text-to-image diffusion models. This new scheme helps reduce overlaps between different objects, allowing the diffusion models to generate multiple elements that are more accurate and faithful to the original prompt. In this figure, we present the results of two different seeds (rows) generated by several baselines and our method. Our method improved the layout, and the images it generates are more faithful to the prompt.  }\label{fig:illustration}
% \end{figure}
% \begin{table*}[htb!]
% \caption{}
% \centering
% \adjustbox{max width=1.3\textwidth}{
% \begin{tabular}{lcc}
% \toprule
%  Symbol  &Description \\
% \midrule
% $t\in (T,0)$&Step index in the diffusion process \\
% $z_t\in \mathbb{R}^{4xhxw}$&Latent map at step $t$ \\
% \bottomrule
% \end{tabular}
% }
%\end{table*}

\section{Related Work}
\label{sec:related}
The groundbreaking potential of generative AI has been unleashed by cutting-edge foundation models, such as Stable Diffusion \cite{rombach2022high}, Imagen \cite{saharia2022photorealistic}, Midjourney, and DALL-E. They serve as the backbone for various generative AI applications, which include image, video, and 3D object generation. These models typically operate in a latent space, mapping from a low-dimensional latent representation to high-dimensional images.
Diffusion models have recently gained attention for their ability to generate high-quality images by iteratively denoising a noise-perturbed input image. Notable work includes the Stable Diffusion model \cite{rombach2022high}, which demonstrated impressive results in generating diverse and realistic images. Diffusion models offer advantages, such as stable training dynamics and the ability to generate high-resolution images, making them a promising choice for image synthesis tasks. The accuracy of the generated image is determined by its adherence to the user-provided text prompt. While existing generative models excel at synthesizing single-object images, generating images with multiple entities poses a more substantial challenge. 

Previous works have attempted to enhance the accuracy of text-to-image diffusion models by improving their semantic faithfulness. One of the major challenges in this regard has been aligning the image outputs with the input prompts, which has been thoroughly discussed in the paper by Tang et al. \cite{tang2022daam}. To address this issue, Kim et al. \cite{liu2022compositional} introduced ComposableDiffusion, a method that allows users to incorporate conjunction and negation operators in prompts. This approach improves concept composition guidance. Further advancements were made with the development of StructureDiffusion \cite{feng2022training}, which suggests segmenting the prompts into noun phrases for exact attention distribution. Wu et al. \cite{wu2023harnessing} proposed a method for controlling the XAMs with a spatial layout generation predictor. Agarwal et al. \cite{agarwal2023star} proposed A-star to minimize concept overlap and change in attention maps through iterations. Kim et al. \cite{kim2023dense} introduced DenseDiffusion, a method for region-specific textual feature accumulation.

The recent paper in \cite{chefer2023attend} aimed to improve attention to neglected tokens, while \cite{li2023divide} suggested two distinct customized objective functions to handle the issues of missing objects and incorrect attribute binding separately. Although these approaches make an effort to resolve the mentioned problems, they still fail multiple object generation. MultiDiffusion \cite{bar2023multidiffusion} combines multiple diffusion generation processes for each token, which helps overcome multi-subject generation difficulties. While MultiDiffusion \cite{bar2023multidiffusion} requires a bounding box for each object in the prompt, which is a significant advantage over our approach and other baselines, it might be capable of overcoming hurdles associated with multi-subject generation. The Attend-and-Excite (A\&E) method \cite{chefer2023attend} focuses solely on neglected objects, but struggles to address the problem effectively when the areas of maximum attention are close. Conversely, Divide-and-Bind \cite{li2023divide} offers an approach to resolve the issue of incorrect attribute.
%  A common generative AI architecture is the generative adverserial network (GAN). Focused on generating visually appealing samples by training a generator network to produce images that are indistinguishable from real ones according to a discriminator network. Another common choice is a variational autoencoders (VAE), which learn a latent space that captures the underlying distribution of the training data, enabling probabilistic generation of images. 
% {\color{red}  lets try to make the related material more precise}

%papers to mention:\cite{meral2023conform}\cite{agarwal2023star} %those without code

\section{Method}

The stable diffusion process is an iterative procedure with indices from $t=T$ to $t=0$, where $T$ is the total number of steps. At each step, a randomly sampled latent map $z_T$ is provided as input to the UNet, which predicts the noise estimation and gradually removes it until it produces the clean version of the latents $z_0$.

The latent map $z_t\in \mathbb{R}^{4\times h\times w}$ represents the output at step $t$. The early steps are important for creating the layout, while the later steps improve the local structure. To condition the text, a prompt $\mathcal{P}$ consisting of $N$ tokens is used, and the latent image $z_t$ is viewed as a grid of $P \times P$ patches. To link each patch to each prompt token, a XAM $A_t\in \mathbb{R}^{P\times P\times N}$ is used. Let $S$ be the set of all subjects in the prompt, and $s$ be the token index. At step $t$, $A_t^s\in \mathbb{R}^{P\times P}$ is the XAM of token $s$ that links each patch to token $s$. %Token $s$ will be considered ``excited" if its maximum attention value is significantly greater than $\frac{1}{N}$, which is calculated by taking the maximum value of $A_t^s$ across all patches.

To synthesize an image that accurately represents a particular subject, several factors must be considered. First, each subject $s$ should have at least one patch with a high value in $A_t^s$, especially during the initial stages of the generation process. Second, the overlap of the attention maps of the subject with the other attention maps should be minimal. %We use a dot product to estimated this overlap $O_{s,s'} := A_t^s \cdot A_t^{s'}$ (the dot product sums across both spatial dimensions).%taken after vectorizing the attention matrices. %\sum_{i,j=1}^{P} (A_t^{s} \odot A_t^{s'})_{ij}$.

% to introduction - Our approach is centered around the concept of generative semantic nursing. We utilize a technique where we progressively optimize the input latent $z_t$ at different steps in order to produce a generation that is less prone to neglecting important aspects and more semantically aligned with the input text .
Our approach contains three steps:
\begin{enumerate}
    \item \textbf{Excite and distinguish:} Occurs in the first $\tau$ steps: $t\in(T,T-\tau)$, where $T>\tau$. This step gently compels all tokens $s\in S$ be excited, i.e., to have a maximal attention score that is high enough, and to spatially separate these maps as much as possible.
    \item \textbf{Rearrange the generation grid:} Occurs once at the end of the first phase. %between phases at step: $t=\tau$. 
    For each subject, extract for each $s\in S$ spatial mask from the XAMs, and then spatially rearrange the masks to minimize their overlap, obtaining the desired masks $M_s$. %Also, relocate pixels in the latents $Z$ based on placement.
    \item \textbf{Follow the masks:} Occurs during all subsequent diffusion steps: $t\in(T-\tau,0)$ and utilizes the masks from the previous phase to guide the spatial arrangement in the XAMs.
\end{enumerate}

\begin{figure}[t]
    \centering
    \includegraphics[width=1.05\linewidth]{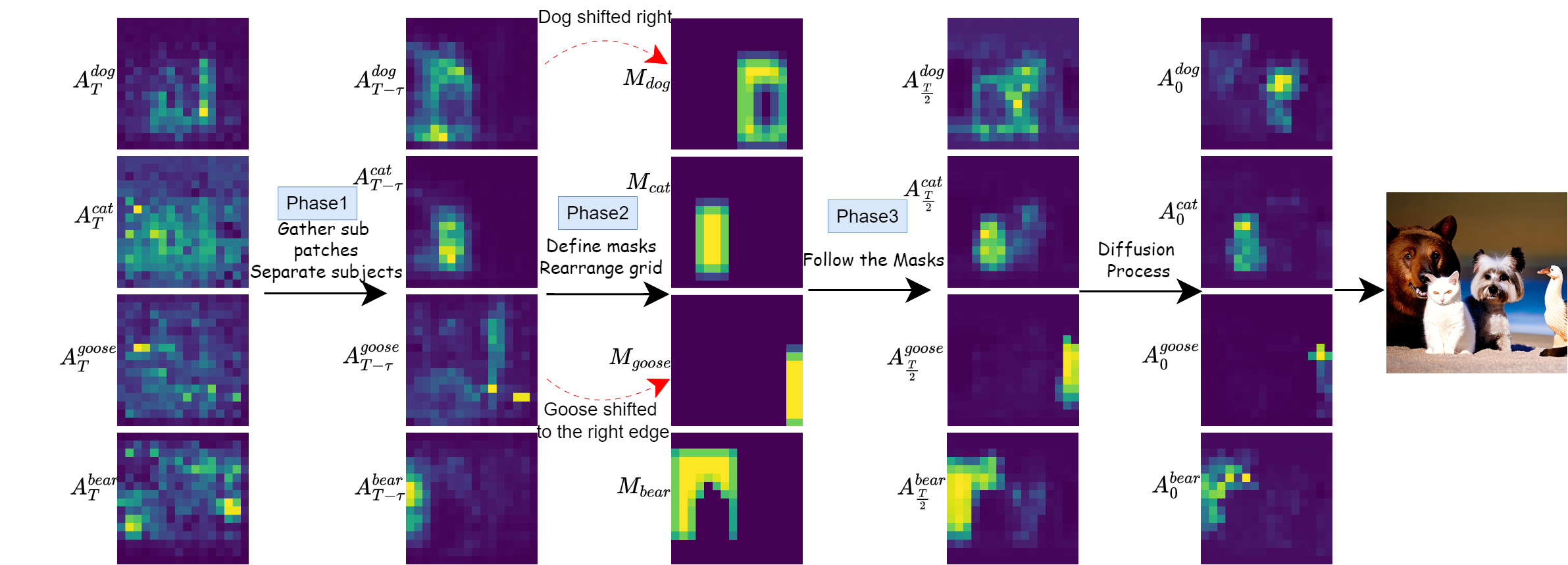} 
\caption{Illustrated herein is the sequential evolution of XAMs throughout the  generation (backward) process. Commencing on the left with $t=T$, the XAMs exhibit a high degree of spatial entropy, signifying an unorganized state. During Phase 1, spanning $t=(T,T-\tau)$, the process strategically consolidates patches pertaining to identical subjects while concurrently segregating the XAMs of distinct subjects. The resulting XAMs at $t=\tau$ manifest enhanced organization and concentrated focus, enabling a preliminary prediction of the subjects' potential generation loci. Phase 2 involves optimizing the spatial arrangement and generating masks that will be used in Phase 3, the masks presented are after Gaussian smoothing. In Phase 3, the attention maps are subtly coerced to align with predefined masks. The extreme right column depicts the remainder of the diffusion process, which is instrumental in mitigating artifacts induced by the optimization process.}
    \label{fig:onemap_highlevel}
\end{figure}

Figure~\ref{fig:onemap_highlevel} illustrates the dynamic interplay of the three phases in shaping the XAMs, thereby steering the layout of image generation towards superior multi-subject generation. 

\noindent\textbf{Phase I\quad} 
It is widely recognized \cite{hertz2022prompt,voynov2023sketch} that the first few diffusion steps, largely determine the layout of the generated image. Therefore, if the goal is to separate the XAMs of subjects, it must be done within these initial steps. Unfortunately, during these early generation steps, it is almost impossible to understand the subject's location in the XAMs. 

As a starting point, consider the A\&E \cite{chefer2023attend} objective, which considers the maximal attention each token $s\in S$ receives and attempts to minimize their neglect by maximizing the attention of the least attended token:
\begin{equation}
L_{A\&E} = \max_{s \in S}  (1 - \max(A^s_{t}))\,, 
\end{equation}
where the second maximization is over all spatial locations of the XAM.

This objective can lead to subject-blending, since the same spatial location can have a high cross-attention score to several tokens. In order to prevent this from occurring, we sort the subject tokens $s$ by their maximal cross-attention values $\max(A^s_{t})$ in a descending manner.

Let $s_0,s_1,s_2,...$ be the list of sorted tokens at time $t$. We define $B_t^{s_0}$ to be the map that is zero everywhere except for a fixed-sized rectangle around the maximal values of $A_t^{s_0}$. We then consider the filtered XAM $\tilde A_t^{s_1} = A_t^{s_1} \odot (1 - B_t^{s_0})$, where $\odot$ marks an element-wise multiplication. We identify the maximal value in the filtered XAM, i.e., the maximal value that is at a distance from the maximal attention value in $A_t^{s_0}$. Let $B$ be a binary mask around the point of highest attention in $\tilde A_t^{s_1}$. We define $B_t^{s_1}$ to be the union (maximum value) of $B_t^{s_0}$ and $B$. 

\begin{figure}[t]
    \centering
    \subfloat[$A_t^{s_0}$]{\includegraphics[width=0.21\linewidth]{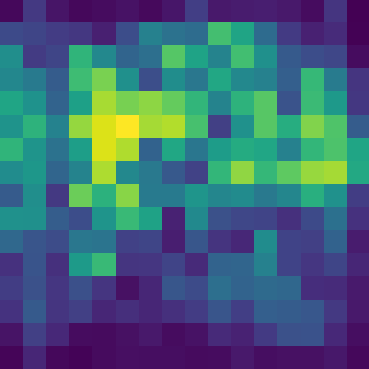}\label{fig:aimg1}}\hspace{0.02\linewidth}
    \subfloat[$A_t^{s_1}$]{\includegraphics[width=0.21\linewidth]{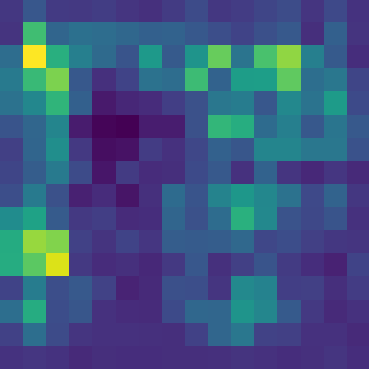}\label{fig:aaimg2}}\hspace{0.02\linewidth}
    \subfloat[$A_t^{s_2}$]{\includegraphics[width=0.21\linewidth]{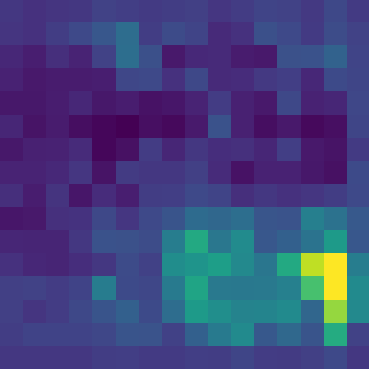}\label{fig:img3}}\hspace{0.02\linewidth}
    \subfloat[$A_t^{s_3}$]{\includegraphics[width=0.21\linewidth]{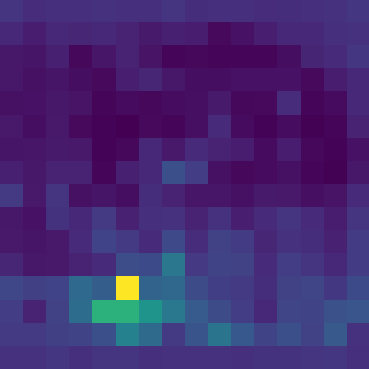}\label{fig:img4}}\\
    \subfloat[$1-B_t^{s_0}$]{\includegraphics[width=0.21\linewidth]{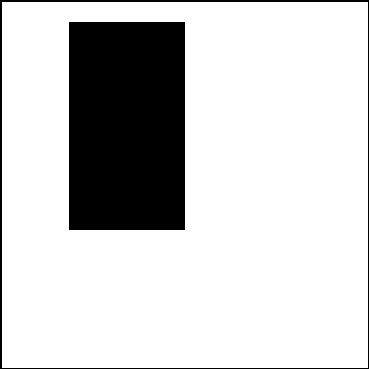}}\hspace{0.02\linewidth}
    \subfloat[$1-B_t^{s_1}$]{\includegraphics[width=0.21\linewidth]{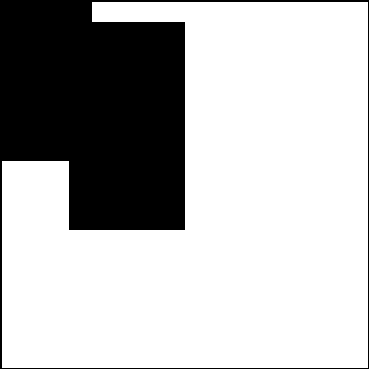}}\hspace{0.02\linewidth}
    \subfloat[$1-B_t^{s_2}$]{\includegraphics[width=0.21\linewidth]{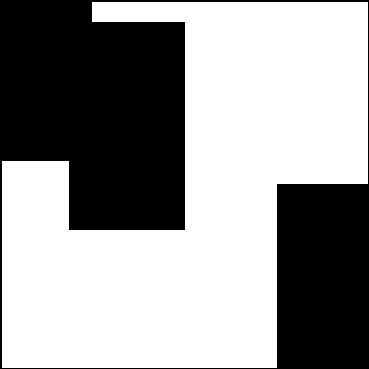}}\hspace{0.02\linewidth}
    \subfloat[$1-B_t^{s_3}$]{\includegraphics[width=0.21\linewidth]{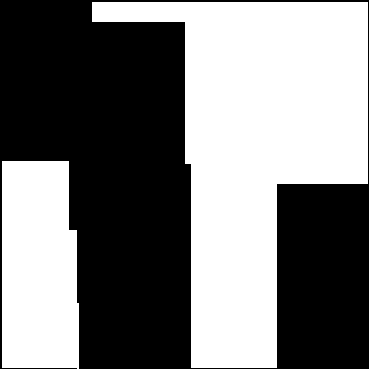}}\hspace{0.02\linewidth}

\caption{The computation of the blocking masks $B_t^s$. The subject tokens $s_i$ are sorted from largest excitation to smallest. At every step, the mask $B_t^{s_i}$ accumulates the masked regions from its predecessors and adds a rectangle around the location of the maximal value in $\tilde A_t^{s} = (1-B_t^{s_{i-1}})\odot A_t^{s_i}$. Note (h) will not be used.}
    \label{fig:B_t_s}
\end{figure}

The process repeats, each time accumulating more regions in which the next token's original attention is masked out, see Figure ~\ref{fig:B_t_s} for an illustration.

The refined A\&E loss term we employ includes these spatial considerations and is given as:
\begin{equation}
L_{B\& E} = \max_{s_i \in S} (1 - \max((1-B^{s_{i-1}}_t)\odot A^s_{t}))\,,    
\end{equation}
where $B_t^{s_{-1}} = 0$ and the second maximization is across all spatial locations as above.

%Figure ~\ref{fig:B_t_s} illustrates the idea behind $(1-B^s_t)$ and $A^s_{t}$ terms.

%{\color{red}WH"Y ADD ANOTHER TERM? MOTIVATE} To mitigate subject overlap in our methodology, we introduce a secondary loss term termed the "overlap loss.'' This loss is formulated as follows:

This loss term $L_{B\& E}$ is effective in having each XAM develop a high value that is spatially far from the high values of other XAMs. However, it does not prevent much of the high values of one XAM to overlap those of other XAMs. To mitigate this, we add a direct loss term per subject token $s$
\begin{equation}
L^s_{ol} = \frac{1}{|S|-1} \sum_{s' \in S\setminus s} \braket{ \bar{A}_t^s ,\bar{A}_t^{s'}}_F ,
\end{equation}
where $\bar{A}_t^s$ represents a dilated version (with a 3x3 kernel) of $A_t^s$,$\braket{}_F$ is the fourbinous inner product
 the sum over $p_i,p_j$ sums across both spatial dimensions. The dilation is applied to prevent the diffusion process from erroneously locating subjects within small gaps or holes that may be present in the image grid. 

Lastly, diffusion models may generate large subjects, leading to multiple subject generation failure due to lack of grid space.  To address this issue, we employ conditional norm regularization, defined as follows:
\begin{equation}
    L^s_{norm} =  [\|A_t^s\|_F>C]\|A_t^s\|_F,
\end{equation}
where $C=\frac{P^2}{|S|}$ accounts for the size of the image as well as the number of subjects. 
%{\color{Red}THIS IS ALL VERY CONFUSING SINCE B\&E IS FOR ALL SUBJECT AND THIS IS PER SUBJECT AND IT IS ALSO NOT CLEAR WHEN YOU OPTIMIZE THIS} 
The overall loss term in the first phase is:
\begin{equation}
   L_{I} = \lambda_{B\& E}L_{B\& E} + \sum_{s\in S} \lambda_{ol} L^s_{ol}+\lambda_{norm}  L^s_{norm} .  
\end{equation}

Where $\lambda_{B\& E}, \lambda_{ol}, \lambda_{norm}$ are scaling factors.

\smallskip 
\noindent\textbf{Phase II\quad} After the first $\tau$ steps, phase II generates a set of per-subject masks $M_s$. It also rearranges the latent space $z_\tau$ to be compatible with the new masks.

%In this phase, assuming that all attention maps have been activated, the goal is to predict which patches belong to subject $s$. It's crucial to acknowledge that a patch $p$ may encompass information pertaining to more than one subject.
The initial masks $\bar{M}^s$ are obtained by thresholding the XAMs at step $t=\tau$. We use the following equation to define our masks: 
%Initially, we assign a binary mask $\bar{M}^s$ for each attention map $A_t^s$ that that depicts the subject spatial location. This mask is generated using a heuristic approach based on the formula:
\begin{equation}
\bar{M}^s[i,j]= 
    \begin{cases}
        1 & A_\tau^s > \gamma \max(A_\tau^s),\\
        0 & \text{otherwise} 
\end{cases}
\end{equation} 

 Where $i,j$ are the spatial indexes. The value of $\gamma$ is set to ensure that the masks are neither too large nor too small, with the default value being 0.2. It is then adjusted automatically within the range of $\gamma \in (0.2, 0.8)$, see supplementary material for the details. %For more information, please refer to the appendix.

%\noind\textbf{Rearranging the grid} 
We next rearrange the masks. We have learned that it is best to move as few objects as possible, and that an object should never be moved above its current location, otherwise, nonphysical scenarios are obtained, which SD cannot generate faithfully.

\begin{figure}[t]
    \centering
{\includegraphics[width=0.85\linewidth]{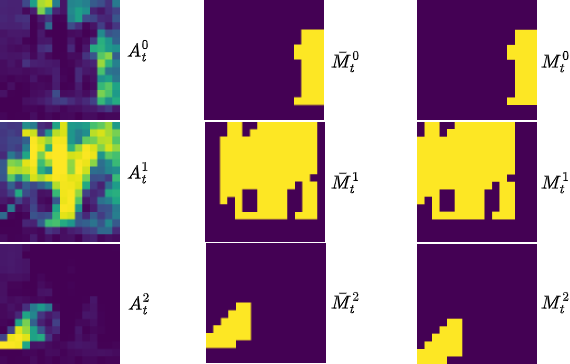}\label{fig:img1}}\hspace{0.02\linewidth}
\caption{This figure illustrates Phase2 of the process. On the left, three masks represent the three subjects. In the middle column, we observe the initial masks, which estimate the patches in the attention maps contributing to each subject. On the right, we observe the final masks, after they have been shifted to their new locations. These final masks will subsequently guide the shifting of attention maps $A_t^s$ towards their new location. }
    \label{fig:phase2}
\end{figure}

The two objects that are moved are chosen based on the size ratio between the overlapping regions with other object over their own size is the highest. The latter is simply the sum of the mask $\bar M^s$. The former is the sum (over all image locations) of the intersection of all overlaps of the form $\bar M^s \cap \bar M^{s'}$ between the initial mask of subject $s$ and that of $s'$.

To determine the new location of a subject $s$, we shift the mask $\bar M^s$ spatially, each time computing the total overlap as above, and selecting the shift that minimizes this quantity. The final set of masks is denoted $M^s$. Please refer to Figure~\ref{fig:phase2} for more intuition.
%. First, we separate the two largest objects. Second, we select $m$ masks with the highest ratio of $\frac{overlap}{size}$. We then determine the new destination of the $m$ masks by performing a naive search over all pixels in the grid $G_s=\sum_{s'\in S\setminus s}M^{s'}$, that do not currently belong to any other $s'\in S\setminus s$, i.e where $G_s$ equal to zero. Although this approach has a complexity of $\mathcal{O}(P^2)$, the value of $P$ is usually small (typically 16/32), and since most of the grid is already occupied by other masks, the search space is negligible. Finally, for each selected mask $M^s$, we identify grid locations that minimize overlap to determine its new destination, the transformation from $\bar{M}_s$ to $M_s$ will be denoted as $ M_s[i,j] = \bar{M}_s[\mathcal{T}([i,j])]  $. For details and discussion on more rigorous solutions, please refer to Section~\ref{subsec:full_def__loss}.

%\noindent\textbf{Patch relocation in the latent map\quad} 
To ease subject reallocation, we rearrange the patches in $z_{\tau}$. Since we know the shift each of the objects we moved has (the result of the above 2D search), we first shift the corresponding locations in $z_{\tau}$. We note that the size of the masks is four times smaller in each dimension than the size of the latent, thus we used an upscaled version of $\bar{M}_s$. 

Shifting requires copying the representation in a certain spatial location to a new location. The original location can remain vacant if no other pixel is copied there. In this case, we use one of two imputation techniques: (1) if there is no background in the prompt, we impute a random vector from the normal distribution, else (2) we copy the latent activation of a the top-k locations from the background token $s_\text{bkg}$ XAM $A_{T-\tau}^{s_\text{bkg}}$.

\begin{figure}[htb]
\begin{tabular}{cccc}
  \raisebox{0.195\height}{\includegraphics[width=.25\linewidth]{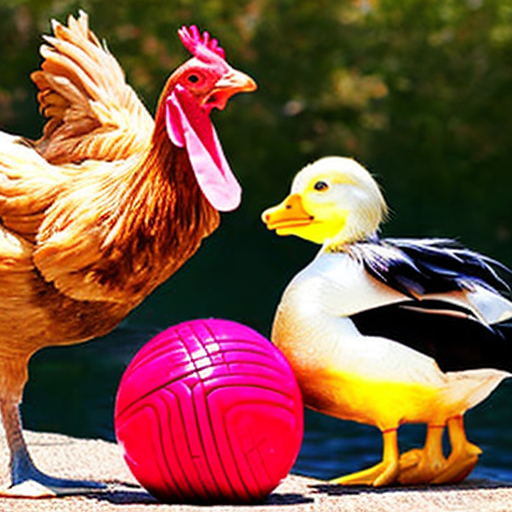}} &
 \includegraphics[width=.25\linewidth]{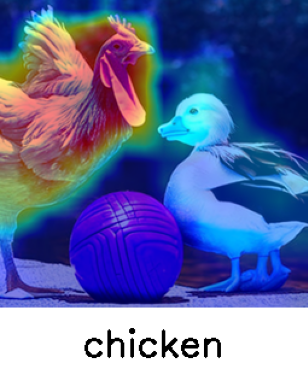}
 \includegraphics[width=.25\linewidth]{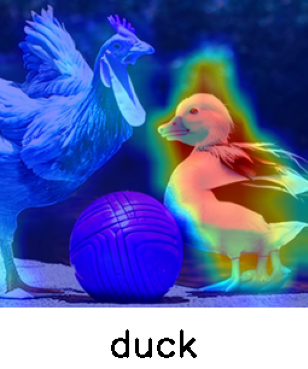}
 \includegraphics[width=.25\linewidth]{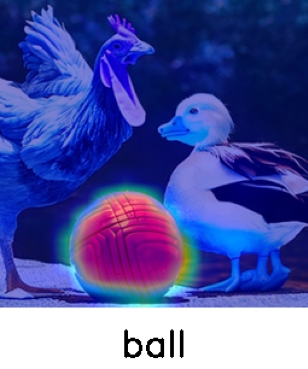} \\

  \raisebox{0.195\height}{\includegraphics[width=.25\linewidth]{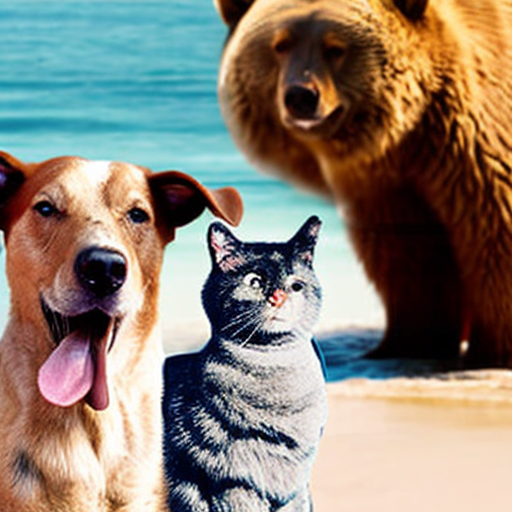}} &
 \includegraphics[width=.25\linewidth]{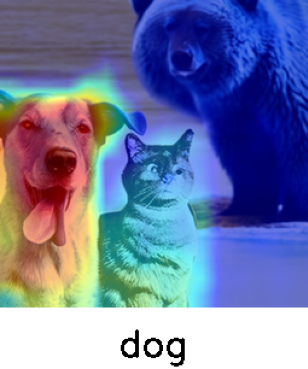}
 \includegraphics[width=.25\linewidth]{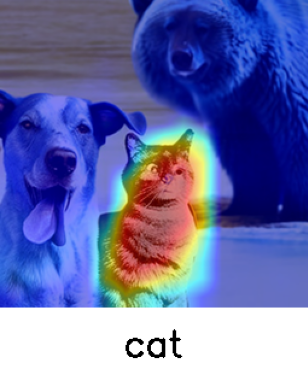}
 \includegraphics[width=.25\linewidth]{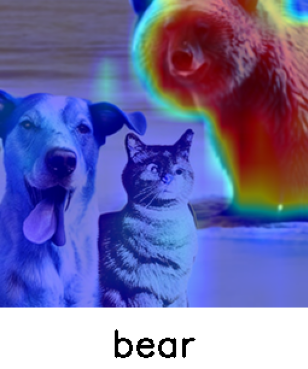} \\
\end{tabular}
\caption{This figure displays the output generated by our method in response to the prompt: "A \textbf{chicken} and a \textbf{duck} with a \textbf{ball} at the beach." and "A \textbf{dog}, a \textbf{cat}, and a \textbf{bear} at the beach". The three rightmost images depict the attention maps at step 9, a pivotal moment in the generation process that significantly influences the layout of the generated image. This step was specifically chosen to highlight its critical role in determining the spatial arrangement of the depicted entities. This visualization helps us analyze the importance of separating the attention maps.}\label{fig:fig_with_attentionmaps1}
\end{figure}

\smallskip 
\noindent\textbf{Phase III\quad}
In this phase of the generation, we encourage $A_t^s$ to follow the masks $M^s$. This is done using two objectives. The first objective keeps the object inside its mask, and it is formulated as:
\begin{equation}
L_\text{inside} = \frac{1}{|S|} \sum_{s\in S}(1-\frac{\braket{A_t^s, M^s}_F}{\|A_t^s\|_1})^2.
\end{equation}
%(the dot product sums across both spatial dimensions). 
The second objective is to try and have the subject's XAM fill the object's mask:% and is formulated as follows:
\begin{equation}
    \quad L_{fill} = \frac{1}{|S|} \sum_{s\in S}(1-\frac{\braket{A_t^s , M^s}_F}{\|M^s \|_1})^2.
\end{equation}
% The third and final objective aims to keep topk image patchs inside the mask of the subject, excited as possible and is formulated as follows:
% \begin{equation}
%     l^s_{top} = topk(A_t^s M^s)
%     \quad L_{top} = \frac{1}{|S|} \sum_{s\in S}(1-l^s_{top})^2
% \end{equation}
The loss term of this phase combines both:
\begin{equation}
     L_{III} = \lambda_{inside}L_{inside}+\lambda_{fill}L_{fill},
\end{equation}
Where $\lambda_{inside}, \lambda_{fill}$ are scaling factors.
%Gotttttt to this point rest is draft
%A\&E \cite{chefer2023attend} encourage token $s$ to be excited by optimizing the latent maps $z_t$: 
%\begin{equation}
%z_t' \leftarrow z_t -\alpha_t\nabla_{z_t}L
%\end{equation}
%Where $L$ is defined as:
%$\begin{equation}
%L = \max_{s \in S} L_s \quad \text{where} \quad L_s = 1 - \max(A^s_{t})    
%end{equation}
 %$ \forall s' \in S \setminus s$
 % this will aid us in defining a rough placement grid $G\in \mathbb{R}^{P\times P}$ which provides the general layout, and is formulated as $G \triangleq \sum_{s\in S}A_t^s$

\section{Experiments}

% \begin{figure}[t]
%     \centering
%     \includegraphics[width=1.02\linewidth]{gallery/ECCV2.drawio.png}
%     \caption{We present a training-free procedure to automatically re-arrange the layout of attention maps in text-to-image diffusion models. This new scheme helps reduce overlaps between different objects, allowing the diffusion models to generate multiple elements that are more accurate and faithful to the original prompt.  }\label{fig:illustration2}
% \end{figure}

\begin{table*}[t]
\caption{This table presents the evaluation results on three test sets comprising only animals. Performance metrics were assessed using the Llava1.5 and Qwen models.}
\label{tab1}
\centering
\adjustbox{max width=1\textwidth}{
\begin{tabular}{lccc|ccc|ccc}
\toprule
Method & \multicolumn{3}{c|}{Two animals} & \multicolumn{3}{c|}{Three animals} & \multicolumn{3}{c}{ Four animals} \\
\cmidrule(r){2-4} \cmidrule(lr){5-7} \cmidrule(l){8-10}
& c-score & l-score & q-score & c-score& l-score & q-score & c-score& l-score & q-score \\
\midrule
Stable Diffusion \cite{rombach2022high} &0.79& 0.83 & 0.64 &0.59& 0.63 & 0.60 &0.35& 0.55 & 0.53  \\
CompDiffusion \cite{liu2022compositional} &0.81& 0.84 & 0.79 &0.79&0.75  & 0.72 &0.61& 0.61 & 0.63 \\
Divide \& Bind \cite{li2023divide}&0.76& 0.92 & 0.92 &0.65& 0.84 & 0.83 &0.57& 0.76 & 0.69 \\
Attend \& Excite \cite{chefer2023attend} &0.82& 0.92 & 0.90 &0.81& 0.79 & 0.73 &0.72& 0.71 & 0.64 \\
SDXL \cite{podell2023sdxl} &0.62& 0.94 & 0.90 &0.45& 0.83 & 0.77 &0.15& 0.80 & 0.65 \\
MultiDiffusion \cite{bar2023multidiffusion} &0.72& 0.92 & 0.92 &0.76& 0.73 &0.70 &0.39 & 0.57 & 0.50 \\
 \rowcolor[HTML]{e8fced} Ours & \bf {0.92} & \bf 0.97& \bf 0.97 &\bf 0.92& \bf 0.90 & \bf 0.91 &\bf 0.85& \bf 0.86 &\bf 0.88 \\
\bottomrule
\end{tabular}
}

\end{table*}

\subsection{Data}
Our experiment's methodology expands the A\&E \cite{chefer2023attend} benchmark by examining text-to-image models in more extreme cases. We use prompts with at least two subjects as templates for the benchmark. These are the templates that were used for the benchmark: (i) “a [animalA] and a [animalB]”, (ii) “a [animalA], a [animalB] and a [animalC]”, (iii) “a [animalA], a [animalB], a [animalC] and a [animalD]”, (iv) “a [animalA], a [animalB] and a [object]”, and (v) “a [animalA], a [animalB], a [animalC] and a [object]”. 

Although this work focuses only on ``subject neglection'' and ``subject blending", we want to demonstrate that solving other issues in image synthesis, such as attribute binding, can be easily combined with our method. We utilized only the ``Bind'' component from Divide-and-Bind \cite{li2023divide} and incorporated it to optimize the latents in conjunction with our suggested loss during phase III. We analyzed two sets of text prompts: (1) “a [colorA][objectA] and a [colorB ][objectB]”, as presented in A\&E \cite{chefer2023attend}. (2) “a [colorA][objectA], a [colorB][objectB] and a [colorC][objectC]”. In total, all methods were evaluated using over 700 prompts, each generated using 5 different seeds.

\subsection{Implementation details} %we use PyTorch\cite{}, 
We use the official Stable Diffusion v2.1 text-to-image model. We use default hyperparameters for all models. All experiments were conducted using an Nvidia Tesla v100 32GB GPU.

\subsection{Baselines} We compared our model to several relevant baselines, including Vanilla SD2.1, Attend-and-Excite \cite{chefer2023attend}, Divide-and-Bind \cite{li2023divide}, Composable Diffusion\cite{liu2022compositional} and MultiDiffusion \cite{bar2023multidiffusion}. Evaluating MultiDiffusion \cite{bar2023multidiffusion} on a large-scale benchmark is challenging, as it requires the user to input the bonding box. To ensure a fair evaluation, we provided the model with three sets of masks that were of a standard size and could be used for generation in a fair manner. Finally, we report the best result out of these three sets of masks.
%\newline 

Despite implementing more advanced training methodologies, leveraging larger neural networks and using larger training data, generating multiple subjects remains a challenge. Unfortunately, due to hardware constraints, we were unable to test our method on SDXL \cite{podell2023sdxl}. However, we will use it to demonstrate the persistent challenge of generating multiple subjects. We used the full pipeline of SDXL \cite{podell2023sdxl}, both the base model and the refiner, all done using Diffusers\cite{wolf2019huggingface} package.

\begin{figure}[t]

    \centering
{\includegraphics[width=0.85\linewidth]{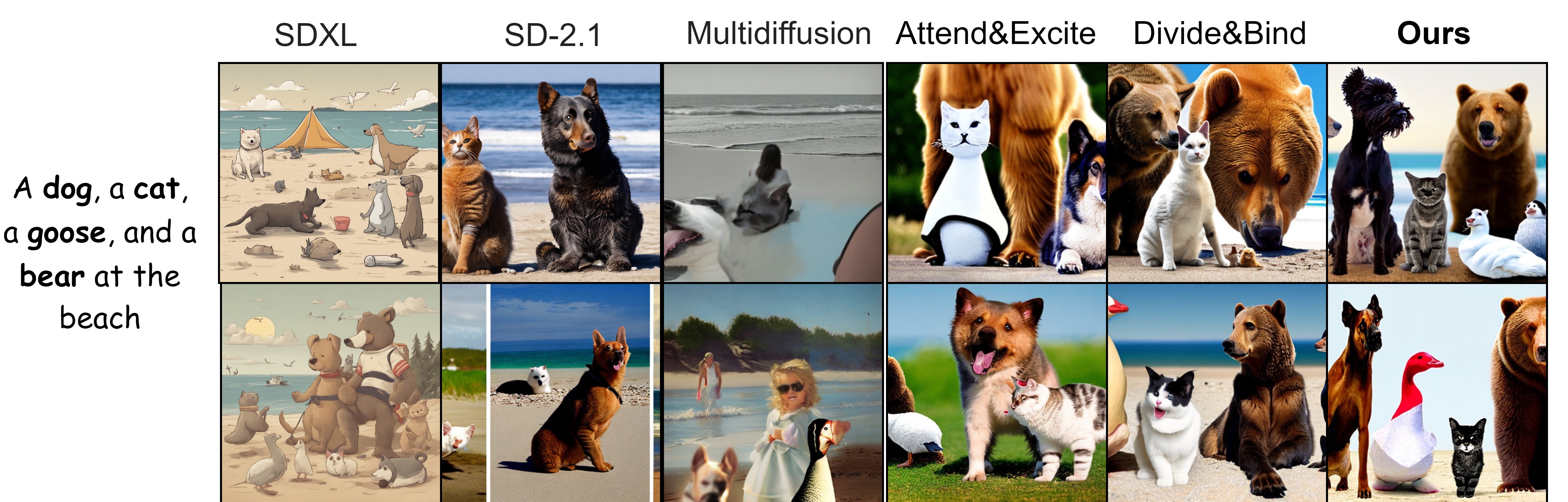} }\hspace{0.02\linewidth}
\caption{This figure shows the outputs of each method examined in this work in response to an input prompt containing four subjects. Other methods suffer from subject blending or are unable to generate all four subjects. In contrast, our method generates an image that is more loyal to the prompt.}
\label{fig:Allbaselines}
\end{figure}

\begin{table*}[t]
\caption{This table shows the results on the set of prompts that consists of subjects including objects.}
\label{tab2}
\centering
\adjustbox{max width=1.3\textwidth}{
\begin{tabular}{lccc|ccc}
\toprule
Method &  \multicolumn{3}{c|}{Two animals and an object} & \multicolumn{3}{c}{ Three animals and an object} \\
\cmidrule(r){2-4} \cmidrule(lr){5-7}
&c-score& l-score & q-score &c-score& l-score & q-score  \\
\midrule
Stable Diffusion \cite{rombach2022high} &0.77& 0.75 & 0.71 &0.59 &0.65 &  0.64  \\
Composable-Diffusion \cite{liu2022compositional} &0.78& 0.75 & 0.73 & 0.75&0.69 & 0.67  \\
Divide \& Bind \cite{li2023divide} &0.70& 0.81 & 0.81 &0.62& 0.72 & 0.70 \\
Attend \& Excite \cite{chefer2023attend} & 0.85&0.86 & 0.82&0.66 & 0.71 & 0.70 \\
SDXL \cite{podell2023sdxl} &0.55& 0.88 & 0.8 &0.46& 0.81 & 0.74 \\
MultiDiffusion \cite{bar2023multidiffusion} &0.63&0.61& 0.58 & 0.61 & 0.58 & 0.57 \\
\rowcolor[HTML]{e8fced} Ours & \bf 0.92& \bf 0.93 &\bf 0.86 &\bf 0.85 &\bf0.87 &\bf0.83 \\
\bottomrule
\end{tabular}}
\end{table*}

\begin{table}[t]
\caption{This table presents the evaluation results on the five test sets evaluated by BLIP2 image-text-matching metric. Each column depicts a different composition of subjects. A=Animal, I=inanimate.}
\label{tab3}
\centering
\begin{tabular}{lcccccc}
\toprule
Method & 2$\times$A & 2$\times$A+I & 3$\times$A & 3$\times$A+I & 4$\times$A \\
\midrule
Stable Diffusion \cite{rombach2022high} & 0.810 & 0.932 & 0.854 & 0.866 & 0.932 \\
Compos-Diffusion \cite{liu2022compositional}  &  0.969&0.972 & 0.955 & 0.976 &0.951\\
Divide\& Bind \cite{li2023divide} & 0.942 & 0.978 & 0.965 & 0.982 & 0.960 \\
Attend \& Excite \cite{chefer2023attend} & 0.957 & 0.985 & 0.968 & 0.991 & 0.982 \\
SDXL \cite{podell2023sdxl} & 0.941 & 0.9275 & 0.934 & 0.990 & 0.982 \\
MultiDiffusion \cite{bar2023multidiffusion} & 0.926 & 0.734 & 0.903 & 0.863 & 0.846 \\
\rowcolor[HTML]{e8fced} Ours &\bf 0.971 & \bf0.991 & \bf0.997 & \bf0.991 & \bf0.998 \\
\bottomrule
\end{tabular}
\end{table}

\begin{table*}[htb!]
\caption{This table presents results for a set of prompts comprising subjects and their associated attribute. It evaluates the model's ability to correctly bind attributes to their respective subjects.}
\label{tab4}
\centering
\adjustbox{max width=1.3\textwidth}{
\begin{tabular}{lccc}
\toprule
Method & 
Two subjects & Three subjects \\
\midrule
Stable Diffusion \cite{rombach2022high} & 0.71 &   0.68 \\
Divide \& Bind \cite{li2023divide} & 0.88&0.83    \\
Attend \& Excite \cite{chefer2023attend} &0.90 & 0.77\\
\rowcolor[HTML]{e8fced} Ours  & {\bf 0.93}&{\bf 0.87}   \\
\bottomrule
\end{tabular}}
\end{table*}

\subsection{Evaluation Metrics}: Thanks to the rapid advancements in the field of vision-language models, we have the capability to assess them automatically. In our study, we utilized Llava1.5 \cite{liu2023llava}  alongside the newly introduced QWEN-VL-Chat \cite{Qwen-VL} with 9.6B parameters, as well as BLIP2\cite{li2023blip}. We used QWEN \cite{Qwen-VL} to describe the animals or objects it identifies, requiring grounding through bounding boxes. This facilitated the creation of a metric, $q$-score, calculated as the ratio of detected subjects to the total subjects in the prompt. For Llava1.5 \cite{liu2023llava}, we prompted the model to identify subject $s$ in the image if it was clearly visible and not blended with the background, for each subject in the prompt. This evaluation, which we term the $l$-score, is comparatively less stringent than the QWEN \cite{Qwen-VL} evaluation and is formulated similarly to the $q$-score.
The full prompt appears in the appendix.
We utilize Llava1.5 \cite{liu2023llava} to extract an additional metric, which will be employed to monitor the numerical quantity of subjects or objects generated by a model. If Llava1.5 \cite{liu2023llava} detects a subject or object, we prompt the model once more, inquiring if there are multiple instances. We refer to this metric as the ``$c$-score", where 1 represents the highest count and 0 indicates the lowest. This score serves not as a simple count but rather as a metric to ensure that the model generates the appropriate number of objects. Its discriminatory power is crucial in distinguishing between models. 

\begin{table*}[t]
\caption{This table presents the evaluation results of the ablation study on two test sets, which include three or four subjects. The first part of the table shows ablation experiments on the objectives in Phase 1. The second part shows the influence of the enhancements components applied in Phase 2, where ``No restart'' means the diffusion process is not restarted after the pixel reallocation. The third block examines the effect of the objectives in Phase 3.}
\label{tab5}
\centering
\adjustbox{max width=1.3\textwidth}{
\begin{tabular}{lccc|ccc}
\toprule
Method &  \multicolumn{3}{c|}{Three subjects} & \multicolumn{3}{c}{ Four subjects} \\
\cmidrule(r){2-4} \cmidrule(lr){5-7}
&c-score& l-score & q-score &c-score& l-score & q-score  \\
\midrule
No $L_{ol}$ &0.91& 0.88 & 0.85 & 0.83&0.84 & 0.83  \\
No $L_{B\&E}$ &0.80& 0.82&0.82 &0.77 & 0.78 & 0.81  \\
No $L_{B\&E}$ \&  $L_{ol}$  & 0.78&0.79 & 0.81&0.71& 0.74 & 0.76 \\
\midrule
No pixel reallocation   &0.89& 0.88 & 0.87 &0.84 &0.83 &  0.86  \\
No restart   &0.91&0.88&0.89&0.84&0.84& 0.88 \\
\midrule
No $L_{fill}$   &0.85&0.85&0.85  & 0.79& 0.78& 0.79 \\
No $L_{inside}$   &0.84&0.85  &0.85  &0.82 &0.80 &0.83    \\
\midrule
Full method  &0.92& 0.90 & 0.91 &0.82& 0.85 & 0.89 \\
\bottomrule
\end{tabular}}
\label{tab:ablation}
\end{table*}
%llava c-score
Additionally, we employed BLIP2 \cite{li2023blip} for image-text matching (ITM), which assesses the likelihood that the given image and text correspond. This process was facilitated using the Lavis package \cite{li-etal-2023-lavis}.

We evaluate the attribute binding experiment using Llava1.5 \cite{liu2023llava}, we prompt the model with two questions: ``Do you see a \{color\} \{obj\} in the image? is it clearly seen?'' and the second is a sequential and harder prompt: ``Do you see a \{obj\} in the image? If \{obj\} exists what is its color?''.

\subsection{Results}

Table~\ref{tab1} and Table~\ref{tab2} present the results on five datasets with a varying number of subjects and objects. In the ``Two animals'' benchmark it is not surprising to see that our method is better by a small gap. Two subjects are what most of the methods improving image generation models attempt and succeed in solving. One can note that as the number of subjects in the prompt increases our method has the lowest decrease in performance, and the gap between the methods increases. Further than that our model keeps its numeric generation stability by keeping a $c-score $ of  0.85 even in the Four subjects benchmark.
MultiDiffusion \cite{bar2023multidiffusion} also suffers from a low $c-score$, which expresses the extent to which the model forcibly generates humans even if they are not needed. 
Attend-and-Excite \cite{chefer2023attend} and Divide-and-Bind \cite{li2023divide} show decent performance along all evaluation metrics.
% \Table~\ref{{tab3}}

Table~\ref{tab3} shows the results on all five benchmarks, using BLIP2 image-text matching metric. While these metrics are more crude (even nonmatching image-prompt pairs often score high if there is some common element), our method is still showing the best performance.

Finally, Table~\ref{tab4} presents the results of the attribute binding experiment, as discussed before, this experiment's goal is to show how well our method can perform by combining methods that tackle different issues.
 As can be seen, our method leveraging only the ``bind''  outperforms all other methods and improves Divide-and-Bind \cite{li2023divide} attribute binding abilities.

\subsubsection{Qualitative Results}

% SDXL in the visual examples (Figure \ref{fig:illustration} and Figure ~\ref{fig:illustration2}) shows an incoherent visual appearance of the animals generated, further, it generates much more characters than required. This overgeneration can be in Table~\ref{tab1} under the $c-score$ column, which is very low for SDXL.

In the next experiment, we illustrate visually the challenges in generating multiple subjects using strong text-to-image diffusion models. Figure ~\ref{fig:opennew} displays a comparison of various baseline models, highlighting their common issues of subject neglection or blending.

For instance, SD2.1 suffers mainly from subject neglection and less from subject blending. Similarly, in the third row, "A gray dog, a brown cat, and a pink bow", the bow's color seems to bleed into the cat, indicating a blending issue. The D\&B suffers from a strong subject blending; for example, in the first row, there is a dog with a frog head; on the right, there is half dog, half cat. A\&E \cite{chefer2023attend} Also suffers from two drawbacks: the first two show a blend between a goat and a duck, and the last row shows no chicken.  It is interesting to see that SDXL \cite{podell2023sdxl} sometimes collapses to "cartoon mode" when the prompt needs too many compositions. SDXL \cite{podell2023sdxl} suffered from subject blending; this can be seen in the second row, where there are ducks with some frog attributes like greenish color and low and wide center of gravity. SDXL also tends to neglect objects, for example, in the first row where the model did not generate a dog and generated some green undefined subject instead of a frog. Our method can better prevent the blending of subjects and enhance the overall performance of subject appearance in the scene. We note that we did not include MultiDiffusion \cite{bar2023multidiffusion} since it was not able to generate reasonable results. 

Figure \ref{fig:fig_with_attentionmaps1} shows the output produced by our method in response to the prompt: "A chicken and a duck with a ball at the beach." and "A dog, a cat, and a bear at the beach." The three images on the right display the attention maps at step 9, a crucial point in the generation process that greatly impacts the layout of the resulting image. We specifically selected this step to emphasize its vital role in shaping the spatial arrangement of the depicted objects. This visualization aids in understanding the significance of individual attention maps.

In Figure  \ref{fig:Allbaselines} we present a qualitative comparison between our method and all baselines. We generated images using the same prompt containing four subjects. Each row shows the generated images for different random seeds. As evidenced by these figures, SD2.1 and MultiDiffusion \cite{bar2023multidiffusion} produce images with blended or incorrect objects. While Attend-and-Excite \cite{chefer2023attend} and Divide-and-Bind \cite{li2023divide} improve image generation quality, they fail to create all four subjects in one image. SDXL \cite{podell2023sdxl} generates cartoonish-style images (even though not requested in the prompt) and also generates images that do not match the set of objects in the prompt. However, our model can create an improved image layout that accommodates all four objects.

\subsection{Searching for a Simpler Solution} 
Recent advancements in large-scale language models (LLMs) have emphasized the importance of input prompts in influencing model outputs. Stable Diffusion online communities consist of users who use and design prompts more frequently than the researchers who developed the models.
We want to embrace their expertise in order to examine much simpler solutions: prompt engineering. We examined three prompts that were suggested in "stable diffusion" related forums for multiple character generation, and we will examine the effect of each prompt on Our method and $A\&E$. 
Table~\ref{tab6} presents the results; one can note that all three prompt variations were not able to create an advantage over the vanilla prompting results presented in Table\ref{tab1}.

\begin{table*}[t]
\caption{This table presents the evaluation results of using prompt engineering for better performance using four and three subject sets. promptA="\{prompt\}, cinematic wide establishing shot", promptB="A photo of \{prompt\} taken from a distance. ", promptC="A photo of \{prompt\} taken from a distance, full body"}
\label{tab6}
\centering
\adjustbox{max width=1.3\textwidth}{
\begin{tabular}{lccc}
\toprule
Method &   \multicolumn{3}{c}{ Four subjects} \\
\cmidrule(r){2-4} 
&c-score& l-score & q-score  \\
\midrule
 SD2.1 promptA    &0.56 &0.63 &  0.59  \\
 A\&E   promptA    &0.64 &0.73 &  0.65  \\
 Ours   promptA   &0.83 &0.84 &  0.86  \\
 \midrule
 SD2.1 promptB    &0.58 &0.63 &  0.59  \\
  A\&E  promptB    &0.67 &0.74 &  0.66  \\
 Ours   promptB    &0.83 &0.82 &  0.85  \\
\midrule
SD2.1 promptC    &0.61 &0.65 &  0.61  \\
 A\&E   promptC    &0.65 &0.73 &  0.65  \\
 Ours   promptC    &0.81 &0.81 &  0.85  \\
\bottomrule
\end{tabular}}
\label{tab:PromptEng}
\end{table*}
\subsection{Ablation Study}

An ablation study was conducted to investigate the effect of removing each component. In our method, we made several  design choices which we examined using an ablation across all phases. In Phase 1, we analyzed $L_{B\& E}$ and $L_{ol}$. In Phase 2, we examined the pixel repositioning in the latents, as well as restarting the diffusion process statistics. Finally, in Phase 3, we checked the impact of No $L_{fill}$ and $L_{inside}$. The results of the ablation study are presented in Table~\ref{tab5}.

The findings suggest that reducing the two losses from phase1 simultaneously results in a significant decrease in performance. When $L_{ol}$ alone is reduced, the effect is less than when $L_{B\&E}$ is reduced, but it still has a substantial impact. The enhancements from Phase 2 have a less pronounced effect on the results, but they still contribute to improved performance.

The objectives of Phase 3 also have a significant impact on performance.

% The results also show that omitting any one of the loss terms in the initial phase, either $L_{ol}$ or $L_{B\& E}$, negatively affects the model's performance. Moreover, dropping both of them together leads to a drastic decrease in performance. This observation supports the notion that these loss terms play a unique role in enforcing the separation of subjects' attention masks from each other, highlighting its importance in maintaining model efficacy. 

\begin{figure}[t]
    \centering
    \includegraphics[width=0.99\linewidth]{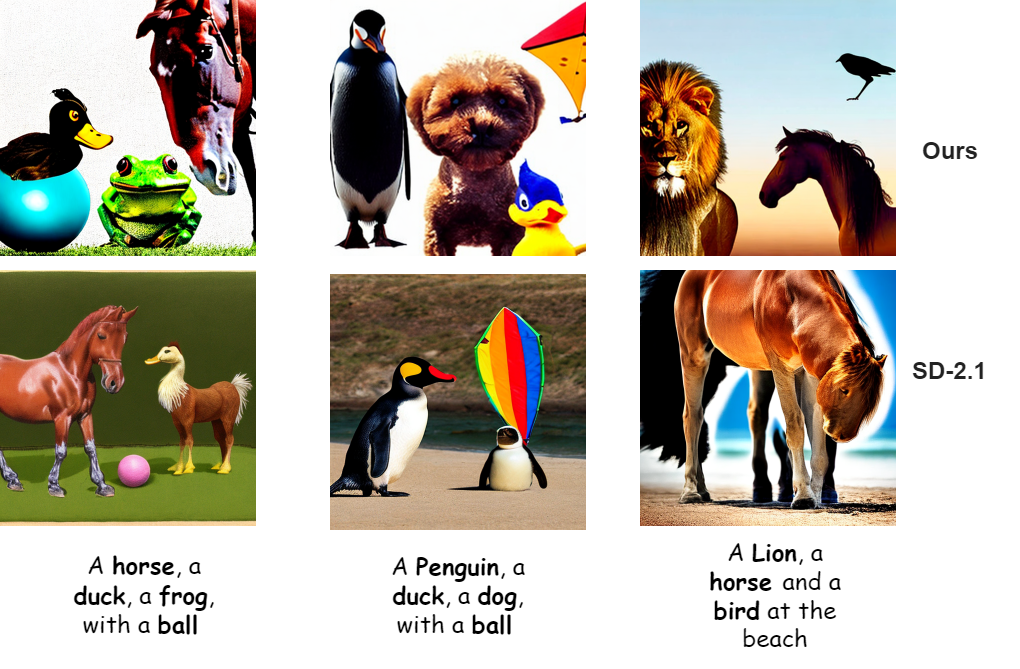}
    \caption{Limitations: the figure presents a few examples of our method and stable diffusion 2.1 generating scenes that feature multiple subjects and objects. Although these cases present multiple objects, their arrangement is unnatural, which may result from imperfect mask generation. Additionally, the image quality may be lacking. Improving the mask generation process is crucial to achieving more successful image generation. To enhance image quality, one can reduce the number of optimization iterations or use a refiner model, such as in the case of SDXL.}\label{fig:open}
\end{figure}

% \begin{table*}[htb!]
% \caption{This table shows the results on the set of prompts that consists of subjects including objects.}
% \centering
% \adjustbox{max width=1.3\textwidth}{
% \begin{tabular}{lcc|cc}
% \toprule
% Method &  \multicolumn{2}{c|}{Three subjects} & \multicolumn{2}{c}{ Four subjects} \\
% \cmidrule(r){2-3} \cmidrule(lr){4-5}
% & l-score & q-score & l-score & q-score  \\
% \midrule
% Stable Diffusion \cite{rombach2022high} & 0.75 & 0.71 & 0.65 &  0.64  \\
% Composable-Diffusion \cite{liu2022compositional} & a & a & a & a  \\
% Divide \& Bind \cite{li2023divide} & 0.78 & 0.74 & 0.68 & 0.65 \\
% Attend \& Excite \cite{chefer2023attend} & 0.86 & 0.82 & 0.77 & 0.69 \\
% MultiDiffusion \cite{bar2023multidiffusion} & a & a & a & a \\
% \bottomrule
% \end{tabular}}
% \end{table*}
%\input{appendix}

\section{Limitations}

There are a few limitations to consider when using our proposed framework. Firstly, it increases the inference time of the synthesis model, although only by a factor of two. However, since rapid generation times are crucial in this field, this limitation is worth taking into account.

Another limitation is that forcing a layout on the diffusion model can impact the image quality negatively. In some cases, the resulting layout may even appear unnatural, as shown in Figure ~\ref{fig:open} for some lower-quality samples. While this effect is usually minor, it does create a tradeoff between staying true to the input prompt and producing high-quality output images.

Lastly, it is worth noting that the method is not aware of subject proportions. As a result, it may generate a small mask for a large subject, and vice versa.

%Of course, such trade-offs are unavoidable. 
\section{Conclusions}
While diffusion models are extremely powerful, they suffer from the same shortcut issue that plagues classifiers and other deep networks~\cite{geirhos2020shortcut,hendrycks2021natural}. In the case of diffusion models, this issue manifests itself as the problem of neglecting and blending subjects and tokens. Naturally, the problem becomes more severe as the prompt becomes increasingly complex. In this work, we offer a multi-step solution that handles multiple aspects of the generation process, including cross-attention and the latent space. The intervention we perform combines pulling the generation process, in inference time, using various loss terms, and more direct editing of the latent space. As we show in an extensive set of experiments, our model enhances the ability of text-to-image diffusion models to generate images with multiple subjects (objects).
Although our method does not attempt to improve the attribute binding phenomena, it is naturally aid this issue. 
The ability to clearly generate multiple subjects and objects is a critical ability of text-to-image models. This capability enables them to create complex scenes effectively.

%\clearpage
\bibliographystyle{IEEEtranN}
\bibliography{main}

% Generated by IEEEtranN.bst, version: 1.14 (2015/08/26)
\begin{thebibliography}{22}
\providecommand{\natexlab}[1]{#1}
\providecommand{\url}[1]{#1}
\csname url@samestyle\endcsname
\providecommand{\newblock}{\relax}
\providecommand{\bibinfo}[2]{#2}
\providecommand{\BIBentrySTDinterwordspacing}{\spaceskip=0pt\relax}
\providecommand{\BIBentryALTinterwordstretchfactor}{4}
\providecommand{\BIBentryALTinterwordspacing}{\spaceskip=\fontdimen2\font plus
\BIBentryALTinterwordstretchfactor\fontdimen3\font minus \fontdimen4\font\relax}
\providecommand{\BIBforeignlanguage}[2]{{%
\expandafter\ifx\csname l@#1\endcsname\relax
\typeout{** WARNING: IEEEtranN.bst: No hyphenation pattern has been}%
\typeout{** loaded for the language `#1'. Using the pattern for}%
\typeout{** the default language instead.}%
\else
\language=\csname l@#1\endcsname
\fi
#2}}
\providecommand{\BIBdecl}{\relax}
\BIBdecl

\bibitem[Rombach et~al.(2022)Rombach, Blattmann, Lorenz, Esser, and Ommer]{rombach2022high}
R.~Rombach, A.~Blattmann, D.~Lorenz, P.~Esser, and B.~Ommer, ``High-resolution image synthesis with latent diffusion models,'' in \emph{Proceedings of the IEEE/CVF conference on computer vision and pattern recognition}, 2022, pp. 10\,684--10\,695.

\bibitem[Saharia et~al.(2022)Saharia, Chan, Saxena, Li, Whang, Denton, Ghasemipour, Gontijo~Lopes, Karagol~Ayan, Salimans, et~al.]{saharia2022photorealistic}
C.~Saharia, W.~Chan, S.~Saxena, L.~Li, J.~Whang, E.~L. Denton, K.~Ghasemipour, R.~Gontijo~Lopes, B.~Karagol~Ayan, T.~Salimans \emph{et~al.}, ``Photorealistic text-to-image diffusion models with deep language understanding,'' \emph{Advances in Neural Information Processing Systems}, vol.~35, pp. 36\,479--36\,494, 2022.

\bibitem[Parmar et~al.(2023)Parmar, Kumar~Singh, Zhang, Li, Lu, and Zhu]{parmar2023zero}
G.~Parmar, K.~Kumar~Singh, R.~Zhang, Y.~Li, J.~Lu, and J.-Y. Zhu, ``Zero-shot image-to-image translation,'' in \emph{ACM SIGGRAPH 2023 Conference Proceedings}, 2023, pp. 1--11.

\bibitem[Tang et~al.(2022)Tang, Liu, Pandey, Jiang, Yang, Kumar, Stenetorp, Lin, and Ture]{tang2022daam}
R.~Tang, L.~Liu, A.~Pandey, Z.~Jiang, G.~Yang, K.~Kumar, P.~Stenetorp, J.~Lin, and F.~Ture, ``What the daam: Interpreting stable diffusion using cross attention,'' \emph{arXiv preprint arXiv:2210.04885}, 2022.

\bibitem[Liu et~al.(2022)Liu, Li, Du, Torralba, and Tenenbaum]{liu2022compositional}
N.~Liu, S.~Li, Y.~Du, A.~Torralba, and J.~B. Tenenbaum, ``Compositional visual generation with composable diffusion models,'' in \emph{European Conference on Computer Vision}.\hskip 1em plus 0.5em minus 0.4em\relax Springer, 2022, pp. 423--439.

\bibitem[Feng et~al.(2022)Feng, He, Fu, Jampani, Akula, Narayana, Basu, Wang, and Wang]{feng2022training}
W.~Feng, X.~He, T.-J. Fu, V.~Jampani, A.~Akula, P.~Narayana, S.~Basu, X.~E. Wang, and W.~Y. Wang, ``Training-free structured diffusion guidance for compositional text-to-image synthesis,'' \emph{arXiv preprint arXiv:2212.05032}, 2022.

\bibitem[Wu et~al.(2023)Wu, Liu, Zhao, Bui, Lin, Zhang, and Chang]{wu2023harnessing}
Q.~Wu, Y.~Liu, H.~Zhao, T.~Bui, Z.~Lin, Y.~Zhang, and S.~Chang, ``Harnessing the spatial-temporal attention of diffusion models for high-fidelity text-to-image synthesis,'' in \emph{Proceedings of the IEEE/CVF International Conference on Computer Vision}, 2023, pp. 7766--7776.

\bibitem[Agarwal et~al.(2023)Agarwal, Karanam, Joseph, Saxena, Goswami, and Srinivasan]{agarwal2023star}
A.~Agarwal, S.~Karanam, K.~Joseph, A.~Saxena, K.~Goswami, and B.~V. Srinivasan, ``A-star: Test-time attention segregation and retention for text-to-image synthesis,'' \emph{arXiv preprint arXiv:2306.14544}, 2023.

\bibitem[Kim et~al.(2023)Kim, Lee, Kim, Ha, and Zhu]{kim2023dense}
Y.~Kim, J.~Lee, J.-H. Kim, J.-W. Ha, and J.-Y. Zhu, ``Dense text-to-image generation with attention modulation,'' in \emph{Proceedings of the IEEE/CVF International Conference on Computer Vision}, 2023, pp. 7701--7711.

\bibitem[Chefer et~al.(2023)Chefer, Alaluf, Vinker, Wolf, and Cohen-Or]{chefer2023attend}
H.~Chefer, Y.~Alaluf, Y.~Vinker, L.~Wolf, and D.~Cohen-Or, ``Attend-and-excite: Attention-based semantic guidance for text-to-image diffusion models,'' \emph{ACM Transactions on Graphics (TOG)}, vol.~42, no.~4, pp. 1--10, 2023.

\bibitem[Li et~al.(2023{\natexlab{a}})Li, Keuper, Zhang, and Khoreva]{li2023divide}
Y.~Li, M.~Keuper, D.~Zhang, and A.~Khoreva, ``Divide \& bind your attention for improved generative semantic nursing,'' \emph{arXiv preprint arXiv:2307.10864}, 2023.

\bibitem[Bar-Tal et~al.(2023)Bar-Tal, Yariv, Lipman, and Dekel]{bar2023multidiffusion}
O.~Bar-Tal, L.~Yariv, Y.~Lipman, and T.~Dekel, ``Multidiffusion: Fusing diffusion paths for controlled image generation,'' 2023.

\bibitem[Hertz et~al.(2022)Hertz, Mokady, Tenenbaum, Aberman, Pritch, and Cohen-Or]{hertz2022prompt}
A.~Hertz, R.~Mokady, J.~Tenenbaum, K.~Aberman, Y.~Pritch, and D.~Cohen-Or, ``Prompt-to-prompt image editing with cross attention control,'' \emph{arXiv preprint arXiv:2208.01626}, 2022.

\bibitem[Voynov et~al.(2023)Voynov, Aberman, and Cohen-Or]{voynov2023sketch}
A.~Voynov, K.~Aberman, and D.~Cohen-Or, ``Sketch-guided text-to-image diffusion models,'' in \emph{ACM SIGGRAPH 2023 Conference Proceedings}, 2023, pp. 1--11.

\bibitem[Podell et~al.(2023)Podell, English, Lacey, Blattmann, Dockhorn, M{\"u}ller, Penna, and Rombach]{podell2023sdxl}
D.~Podell, Z.~English, K.~Lacey, A.~Blattmann, T.~Dockhorn, J.~M{\"u}ller, J.~Penna, and R.~Rombach, ``Sdxl: Improving latent diffusion models for high-resolution image synthesis,'' \emph{arXiv preprint arXiv:2307.01952}, 2023.

\bibitem[Wolf et~al.(2019)Wolf, Debut, Sanh, Chaumond, Delangue, Moi, Cistac, Rault, Louf, Funtowicz, et~al.]{wolf2019huggingface}
T.~Wolf, L.~Debut, V.~Sanh, J.~Chaumond, C.~Delangue, A.~Moi, P.~Cistac, T.~Rault, R.~Louf, M.~Funtowicz \emph{et~al.}, ``Huggingface's transformers: State-of-the-art natural language processing,'' \emph{arXiv preprint arXiv:1910.03771}, 2019.

\bibitem[Liu et~al.(2023)Liu, Li, Wu, and Lee]{liu2023llava}
H.~Liu, C.~Li, Q.~Wu, and Y.~J. Lee, ``Visual instruction tuning,'' in \emph{NeurIPS}, 2023.

\bibitem[Bai et~al.(2023)Bai, Bai, Yang, Wang, Tan, Wang, Lin, Zhou, and Zhou]{Qwen-VL}
J.~Bai, S.~Bai, S.~Yang, S.~Wang, S.~Tan, P.~Wang, J.~Lin, C.~Zhou, and J.~Zhou, ``Qwen-vl: A frontier large vision-language model with versatile abilities,'' \emph{arXiv preprint arXiv:2308.12966}, 2023.

\bibitem[Li et~al.(2023{\natexlab{b}})Li, Li, Savarese, and Hoi]{li2023blip}
J.~Li, D.~Li, S.~Savarese, and S.~Hoi, ``Blip-2: Bootstrapping language-image pre-training with frozen image encoders and large language models,'' \emph{arXiv preprint arXiv:2301.12597}, 2023.

\bibitem[Li et~al.(2023{\natexlab{c}})Li, Li, Le, Wang, Savarese, and Hoi]{li-etal-2023-lavis}
\BIBentryALTinterwordspacing
D.~Li, J.~Li, H.~Le, G.~Wang, S.~Savarese, and S.~C. Hoi, ``{LAVIS}: A one-stop library for language-vision intelligence,'' in \emph{Proceedings of the 61st Annual Meeting of the Association for Computational Linguistics (Volume 3: System Demonstrations)}.\hskip 1em plus 0.5em minus 0.4em\relax Toronto, Canada: Association for Computational Linguistics, Jul. 2023, pp. 31--41. [Online]. Available: \url{https://aclanthology.org/2023.acl-demo.3}
\BIBentrySTDinterwordspacing

\bibitem[Geirhos et~al.(2020)Geirhos, Jacobsen, Michaelis, Zemel, Brendel, Bethge, and Wichmann]{geirhos2020shortcut}
R.~Geirhos, J.-H. Jacobsen, C.~Michaelis, R.~Zemel, W.~Brendel, M.~Bethge, and F.~A. Wichmann, ``Shortcut learning in deep neural networks,'' \emph{Nature Machine Intelligence}, vol.~2, no.~11, pp. 665--673, 2020.

\bibitem[Hendrycks et~al.(2021)Hendrycks, Zhao, Basart, Steinhardt, and Song]{hendrycks2021natural}
D.~Hendrycks, K.~Zhao, S.~Basart, J.~Steinhardt, and D.~Song, ``Natural adversarial examples,'' in \emph{Proceedings of the IEEE/CVF Conference on Computer Vision and Pattern Recognition}, 2021, pp. 15\,262--15\,271.

\end{thebibliography}
\end{document}